\documentclass{article}
\usepackage{ijcai20}

\usepackage{times}

\usepackage{soul}
\usepackage{url}
\usepackage[hidelinks]{hyperref}
\usepackage[utf8]{inputenc}
\usepackage[small]{caption}
\usepackage{graphicx}
\usepackage{amsmath}
\usepackage{amsthm}
\usepackage{amssymb}
\usepackage{booktabs}
\usepackage{subcaption}
\usepackage[tworuled]{algorithm2e}
\usepackage{multirow}

\newcommand{\citet}[1]{\citeauthor{#1} \shortcite{#1}}

\title{Disentangled Representation Learning with Wasserstein Total Correlation}

\author{
Yijun Xiao\and
William Yang Wang\\
\affiliations
University of California, Santa Barbara\\
\emails
\{yijunxiao, william\}@cs.ucsb.edu
}

\begin{document}

\maketitle

\begin{abstract}
Unsupervised learning of disentangled representations involves uncovering of different factors of variations that contribute to the data generation process. Total correlation penalization has been a key component in recent methods towards disentanglement. However, Kullback-Leibler (KL) divergence-based total correlation is metric-agnostic and sensitive to data samples. In this paper, we introduce Wasserstein total correlation in both variational autoencoder and Wasserstein autoencoder settings to learn disentangled latent representations. A critic is adversarially trained along with the main objective to estimate the Wasserstein total correlation term. We discuss the benefits of using Wasserstein distance over KL divergence to measure independence and conduct quantitative and qualitative experiments on several data sets. Moreover, we introduce a new metric to measure disentanglement. We show that the proposed approach has comparable performances on disentanglement with smaller sacrifices in reconstruction abilities.
\end{abstract}

\section{Introduction}
Representation learning makes the assumption that high-dimensional observations are generated (often with noise) from a set of factors and these observations can be effectively represented using dense yet much lower dimensional latent variables. The goal of representation learning, therefore, is to find useful transformations of the observations to the latent space.

Recently, disentangled representation learning attracts more attention \cite{chen2016infogan,chen2018isolating,denton2017unsupervised,dupont2018learning,higgins2017beta,hsu2017unsupervised,locatello2019challenging,kim2018disentangling,kumar2017variational,mathieu2016disentangling,reed2014learning,yang2015weakly}. There are studies focusing on a formal definition of disentangled representations \cite{higgins2018towards}, however there is no widely accepted formal notation yet. Many of the studies adopt the definition from \cite{bengio2013representation}: a change in one dimension corresponds to a change in one factor of variation, while being relatively invariant to changes in other factors. In other words, factorial representations with statistically independent variables are preferred \cite{achille2018emergence,schmidhuber1992learning}.

Many state-of-the-art approaches for unsupervised disentangled representation learning are based on modifications on the variational autoencoder (VAE) \cite{kingma2013auto,rezende2015variational} objective. Specifically, penalization on the total correlation of the latent variable distribution is shown to be an essential ingredient due to its independence encouraging property \cite{chen2018isolating,kim2018disentangling}. \citet{kim2018disentangling} estimate the total correlation using density ratio trick with adversarial training while \citet{chen2018isolating} adopt a tractable but biased Monte Carlo estimator. 

A major challenge in these studies is the estimation of the KL divergence-based total correlation term. Estimating KL divergence in the mini-batch setting is unreliable as it is sensitive to small differences in data samples. KL divergence is also metric-agnostic which means it ignores the geometrical fact that some points in the metric space are closer to others. Two distributions can have arbitrarily large KL divergence even if their samples are very close.

In this paper, we introduce a Wasserstein distance version of total correlation and propose to learn disentangled representations by optimizing over penalized auto-encoding objectives. Furthermore, we introduce a new metric to measure disentanglement motivated by the same limitation of KL divergence on measuring mutual information. Our experiments show that the proposed approach has comparable if not better performances on disentanglement while maintaining minimum sacrifices to the reconstruction ability compared to several baselines.

In summary, our main contributions are: (1) we introduce Wasserstein total correlation, a Wasserstein distance version of total correlation, and apply it to disentangled representation learning; (2) we introduce Wasserstein dependency gap as a new metric to measure disentanglement; (3) we give quantitative comparisons of the proposed approach and several other state-of-art models.

%\section{Related work}

\section{Background}
\subsection{Disentangled representations and total correlation}
Variational autoencoders (VAE) \cite{kingma2013auto,rezende2015variational} are latent variable models that aim to maximize the evidence lower bound (ELBO) of the data likelihood. The loss function of VAE is given as
\begin{align}
\label{equ:vae}
    \mathcal{L_{\mathrm{VAE}}}:= &-\mathbb{E}_{q(z | x) p(x)}[\log p(x | z)]\nonumber\\
    &+\mathbb{E}_{p(x)}[\mathrm{KL}(q(z | x) \| p(z))]
\end{align}
where $x$ represents the observation and $z$ is the latent code. $q(z|x)$ is the approximated posterior distribution of $z$ and $p(x|z)$ represents the generating process.
The $\beta$-VAE \cite{higgins2017beta} attempts to encourage disentangled representations by heavily penalizing the KL divergence term between the posterior and the prior of the latent code $z$. However, it is not obvious why the penalization can lead to latent variables that exhibit disentanglement.

\citet{burgess2018understanding} explain the disentangling effect of $\beta$-VAE from the information bottleneck perspective. Several other studies \cite{chen2018isolating,kim2018disentangling} argue that total correlation (TC) \cite{watanabe1960information} is one of the key reasons behind the disentangling behavior of $\beta$-VAE. In particular, \citet{chen2018isolating} show that the KL divergence term in Equation \ref{equ:vae} can be decomposed into the sum of three terms: index-code mutual information, total correlation, and dimension-wise KL. They propose a tractable but biased Monte Carlo estimator of the total correlation term and penalize it during training. \citet{kim2018disentangling}, on the other hand, use the density-ratio trick \cite{nguyen2010estimating,sugiyama2012density} to estimate the total correlation of the latent variable. Their model consistently underestimates the true TC, yet the gradients obtained are sufficient for encouraging independence in the code distribution.

\citet{kumar2017variational} propose to penalize the mismatch between the aggregated posterior and a factorized prior in order to encourage disentanglement. Interestingly, this essentially becomes a hybrid of variational and Wasserstein autoencoders \cite{ambrogioni2018wasserstein,tolstikhin2018wasserstein}. Although total correlation does not explicitly appear in their formulation, it is involved implicitly the same way as in $\beta$-VAE. 

Most studies use continuous latent variables to model the generating factors, \citet{dupont2018learning} show that a joint model of continuous and discrete latent variables performs better when a discrete generative factor is prominent.

\subsection{Wasserstein distance}

Optimal transport (OT) problem \cite{villani2003topics} introduces a rich class of distance measures between probability distributions. The $p$-\textit{Wasserstein distance} between two probability distributions $P, Q$ with support on the metric space $(\mathcal{X}, d)$ is given as:
\begin{align}
    W_p^p\left(P, Q\right) =\inf _{\gamma \in \Gamma\left(P, Q\right)} \mathbb{E}_{(x, y) \sim \gamma}[d^p(x, y)]
\end{align}
where $p\geq 1$, $\Gamma(P,Q)$ represents the set of distributions on $\mathcal{X}\times\mathcal{X}$ and with marginals $P$ and $Q$ respectively. When $p=1$, we have the \textit{Wasserstein}-$1$ or \textit{Earth-Mover} (EM) distance. The following Kantorovich-Rubinstein duality holds for Wasserstein-1 distance:
\begin{align}
\label{equ:w1}
    W_{1}\left(P, Q\right)=\sup _{f \in \mathcal{F}_{L}} \mathbb{E}_{x \sim P}[f(x)]-\mathbb{E}_{y \sim Q}[f(y)]
\end{align}
where $\mathcal{F}_{L}$ denotes the set of all 1-Lipschitz functions on $(\mathcal{X}, d)$. This allows us to maximize over a constrained function space to solve for the Wasserstein-$1$ distance. Practically, \citet{arjovsky2017wasserstein} propose to solve the following maximization problem instead of Equation \ref{equ:w1}:
\begin{align}
\label{equ:w1_alt}
     W_{1}\left(P, Q\right)\approx \frac{1}{L}\max _{\theta \in \Theta} \mathbb{E}_{x \sim P}\left[f_{\theta}(x)\right]-\mathbb{E}_{y \sim Q}\left[f_{\theta}(y)\right]
\end{align}
where $\{f_\theta\}_{\theta\in\Theta}$ is a family of $\theta$-parameterized functions on $(\mathcal{X}, d)$ that are all $L$-Lipschitz for some $L>0$.

Wasserstein distance is metric-aware and robust to estimation with random samples. It has recently been studied extensively in the context of deep generative models \cite{ambrogioni2018wasserstein,arjovsky2017wasserstein,gulrajani2017improved,ozair2019wasserstein,tolstikhin2018wasserstein} and shows very promising results.

\begin{figure}[t]
\centering
\begin{minipage}[b]{0.3\textwidth}
\centering
\includegraphics[width=\textwidth]{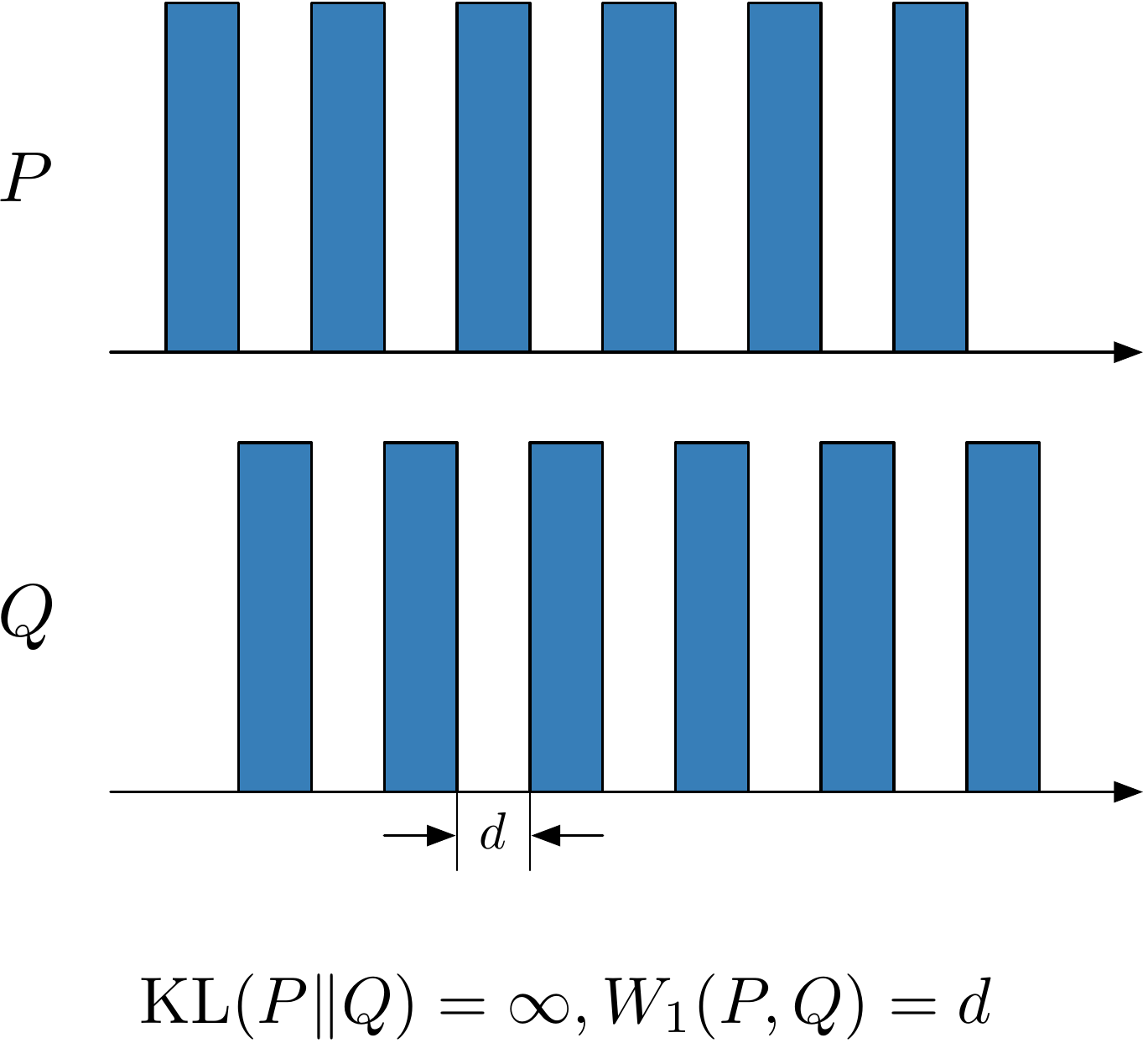}
\\
(a)
\end{minipage}
\hfill
\begin{minipage}[b]{0.45\textwidth}
\centering
\includegraphics[width=\textwidth]{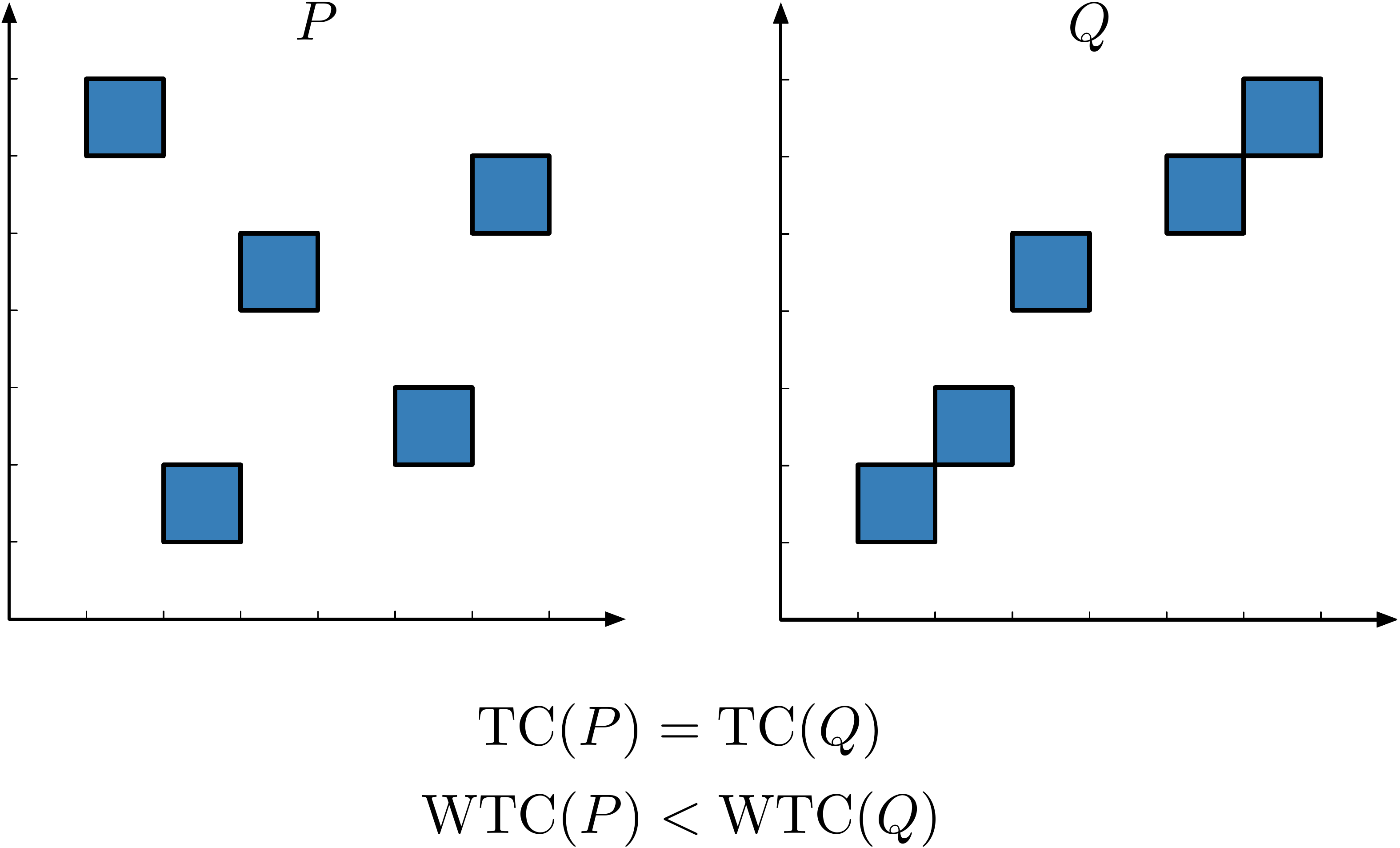}
\\
(b)
\end{minipage}

\caption{(a) One-dimensional example of the difference between KL divergence and Wasserstein distance; (b) two-dimensional example of different measurements from total correlation and Wasserstein total correlation.}
    \label{fig:kld_vs_emd}
\end{figure}

\section{Method}
In this section, we define the Wasserstein total correlation and introduce it in two model settings for disentanglement.
\subsection{Wasserstein total correlation}
Total correlation can be written as the KL divergence between the joint distribution and the product of all marginals. Although it is a good dependency measure among different variable dimensions, it is difficult to estimate in mini-batch training settings. KL divergence is agnostic to the metric of the underlying data distribution and is sensitive to small differences in the data samples. KL divergence can be large even if the underlying data samples are very similar. On the other hand, Wasserstein distance is metric-aware and represents the difference between two distributions in terms of the actual distance between data samples. For example, consider two uni-variate distributions $P, Q$ with disjoint support as shown in Figure \ref{fig:kld_vs_emd}(a). The KL divergence between these two distributions is infinity regardless of $d$. However, the samples from these two distributions become indistinguishable (without knowledge of the ground truth distributions) as $d$ approaches to zero.

Therefore, we propose a new correlation measure as a replacement for the total correlation term used commonly in disentangled representation learning, which we refer to as the \textit{Wasserstein total correlation} (WTC). It is mathematically defined as the Wasserstein-1 distance between the joint distribution and the product of all marginal distributions:
\begin{align}
\label{equ:wtc}
    \text{TC}_\mathcal{W}\left(p(x)\right)=W_1\left(p(x),\prod_ip_i(x_i)\right)
\end{align}
where $x_i$ denotes the $i$-th dimension in $x$ and $p_i$ represents the marginal distribution of $x_i$. This definition is in spirit similar to the Wasserstein dependency measure \cite{ozair2019wasserstein}, but the usage of Wasserstein dependency measure is focused on maximizing the mutual information between two variables. Whereas we seek to minimize WTC as an approach to learn disentangled representations.

Figure \ref{fig:kld_vs_emd}(b) shows a two-dimensional example where WTC better represents independence than TC. $P$ and $Q$ are two-dimensional uniform distributions with support on the disjoint squares. It is easy to check that total correlations of $P$ and $Q$ are equal while WTCs of the two are not. This scenario can potentially happen when calculating the total correlation of latent variables as only a batch of samples is used to approximate the aggregated posterior distributions.

With the definition of the Wasserstein total correlation, we now discuss how to estimate the term with Monte Carlo estimation. 
Firstly, assume $x\in\mathbb{R}^n$ and let $\bar{p}(x)=\prod_{i=1}^np_i(x_i)$ be the product of marginals. By Equation \ref{equ:w1_alt}, \ref{equ:wtc}, we have
\begin{align}
    \text{TC}_\mathcal{W}\left(p(x)\right)\approx\frac{1}{L}\max _{\theta \in \Theta} \mathbb{E}_{x \sim p(x)}\left[f_{\theta}(x)\right]-\mathbb{E}_{\bar{x} \sim \bar{p}(x)}\left[f_{\theta}(\bar{x})\right]
\end{align}

\begin{algorithm}[t]
\DontPrintSemicolon
\IncMargin{1in}
\KwIn{$\{z^{(i)}\in\mathbb{R}^d:i=1,\cdots,B\}$.}
\For{$j=1$ \textbf{to} $d$}{
$\pi\leftarrow$ random permutation on $\{1,\cdots,B\}$\;
$\left(z_{j}^{(i)}\right)_{i=1}^{B} \leftarrow\left(z_{j}^{(\pi(i))}\right)_{i=1}^{B}$
}
\KwOut{$\left\{z^{(i)} : i=1, \dots, B\right\}$}
\caption{\texttt{permute\_dims}}
\label{alg:permute}
\end{algorithm}

Assuming we could sample from $p(x)$, then we can also sample from $\bar{p}(x)$ by generating $n$ samples from $p(x)$ and ignoring all but one dimension for each sample. We may use a more efficient approach by randomly permuting within a sampled batch for each dimension as outlined in Algorithm \ref{alg:permute} \cite{kim2018disentangling} (this is a standard trick used in independence testing literature \cite{arcones1992bootstrap}). A Monte Carlo estimator of the WTC term can be constructed as:
\begin{align}
    \text{TC}_\mathcal{W}\left(p(x)\right)\approx\frac{1}{L}\max _{\theta \in \Theta}\frac{1}{B}\left(\sum_{i=1}^Bf_{\theta}(x^{(i)})-\sum_{i=1}^Bf_{\theta}(\bar{x}^{(i)})\right)
\end{align}
where $\{x^{(i)}\}_{i=1}^B$ are samples from the joint distribution $p(x)$, and $\{\bar{x}^{(i)}\}_{i=1}^B$ are samples from $\bar{p}(x)$ using above described approaches.

A key constraint of the above approximations is that the family of functions we optimize on must be $L$-Lipschitz regardless of $\theta\in\Theta$. We choose to apply gradient penalty \cite{gulrajani2017improved} to enforce the Lipschitz constraint.

In the following two sections, we demonstrate the use of Wasserstein total correlation in the settings of variational and Wasserstein autoencoders.

\subsection{Variational autoencoders with WTC}

\citet{chen2018isolating} shows that the KL divergence term in a VAE can be decomposed into three terms: index-code mutual information, total correlation, and dimension-wise KL. Disentanglement can be achieved by assigning extra weight to the total correlation term. Both FactorVAE \cite{kim2018disentangling} and $\beta$-TCVAE \cite{chen2018isolating} adopt this strategy with different approaches to estimate the total correlation term.

We use a similar strategy to encourage disentanglement, but replace the total correlation term with WTC. The WTC penalized VAE objective can be written as:
\begin{align}
    \mathcal{L_{\mathrm{WTC-VAE}}}:= & -\mathbb{E}_{q(z | x) p(x)}[\log p(x | z)]\nonumber\\
    &+\mathbb{E}_{p(x)}[\mathrm{KL}(q(z | x) \| p(z))]\nonumber\\
    &+\gamma \max _{\theta \in \Theta} \mathbb{E}_{q(z)}\left[f_{\theta}(z)\right]-\mathbb{E}_{\bar{q}(\bar{z})}\left[f_{\theta}(\bar{z})\right]
\end{align}

The training objective function involves three major components: the encoder $q(z|x)$, the decoder $p(x|z)$, and the critic $f_\theta(z)$ used to calculate WTC. The optimization becomes a min-max game between the autoencoder and the critic. The training process is outlined in Algorithm \ref{alg:wtc_vae} without gradient penalty. We could choose to run multiple critic optimizing steps before performing a VAE update step, but in practice a single critic update works reasonably well.

\begin{algorithm}[t]
\DontPrintSemicolon
\IncMargin{1in}
\KwIn{Regularization coefficient $\gamma$, learning rate $\alpha$, batch size $B$.}
\KwIn{initial critic parameter $\theta$, initial VAE parameter $w$}
\While{$(\theta, w)$ not converged}{
sample batch $(x^{(i)})_{i\in\mathcal{B}}$ of size $B$\;
sample $z^{(i)}\sim q_w(z|x^{(i)})$ $\forall i\in \mathcal{B}$\;
$(\bar{z}^{(i)})_{i\in\mathcal{B}}\leftarrow$\texttt{permute\_dims}($(z_w^{(i)})_{i\in\mathcal{B}}$)\;
$\theta\leftarrow\theta+\alpha\cdot\nabla_\theta\frac{1}{B}\sum_{i\in\mathcal{B}}[f_\theta(z^{(i)})-f_\theta(\bar{z}^{(i)})]$\;
$w\leftarrow w-\alpha\cdot\nabla_w\frac{1}{B}\sum_{i\in\mathcal{B}}[\text{KL}(q_w(z|x^{(i)})\|p(z))-\log{p_w(x^{(i)}|z^{(i)})}+\gamma(f_\theta(z^{(i)})-f_\theta(\bar{z}^{(i)}))]$
}
\caption{Training of WTC VAE}
\label{alg:wtc_vae}
\end{algorithm}

\subsection{Wasserstein autoencoders with WTC}
Wasserstein autoencoders (WAEs) \cite{ambrogioni2018wasserstein,tolstikhin2018wasserstein} are alternatives to VAEs to jointly learn a generative model and an inference model on high-dimensional inputs. WAE is a generalization of adversarial auto-encoders \cite{makhzani2015adversarial} and it shares many properties of VAEs while generating samples of better quality \cite{tolstikhin2018wasserstein}. The loss function of a WAE has the following form:
\begin{align}
    \mathcal{L_{\mathrm{WAE}}}:=\mathbb{E}_{q(z | x) p(x)}[c(x,G(z))]+\beta D_z(q(z),p(z))
\end{align}
where $G$ represents the decoder; $c(\cdot,\cdot)$ is any measurable cost function; $D_z$ is an arbitrary divergence measure between two distributions. In \cite{tolstikhin2018wasserstein}, $D_z$ is chosen to be either Jensen–Shannon (JS) divergence or the maximum mean discrepancy (MMD). We could also choose Wasserstein-$1$ distance and resulting in:
\begin{align}
\label{equ:wae}
    \mathcal{L_{\mathrm{WAE}}}:=&\mathbb{E}_{q(z | x) p(x)}[c(x,G(z))]+\beta W_1(q(z),p(z))\nonumber\\
    \leq &\mathbb{E}_{q(z | x) p(x)}[c(x,G(z))]+\beta W_1(\prod_iq_i(z_i),p(z))\nonumber\\
    & +\beta W_1(q(z),\prod_iq_i(z_i))
\end{align}

Equation \ref{equ:wae} uses the fact that Wasserstein-$1$ is a metric. We could then have the definition of the WTC-WAE objective as:
\begin{align}
    \mathcal{L_{\mathrm{WTC-WAE}}}:= & \mathbb{E}_{q(z | x) p(x)}[c(x,G(z))]\nonumber\\
    &+\beta  \max _{\phi \in \Phi} \mathbb{E}_{\bar{q}(\bar{z})}\left[g_{\phi}(z)\right]-\mathbb{E}_{p(z)}\left[g_{\phi}(z)\right] \nonumber\\
    &+\gamma \max _{\theta \in \Theta} \mathbb{E}_{q(z)}\left[f_{\theta}(z)\right]-\mathbb{E}_{\bar{q}(\bar{z})}\left[f_{\theta}(\bar{z})\right]
\end{align}
where $g, f$ are critics for the two different Wasserstein-$1$ distance estimations; $\beta,\gamma>0$ and here we choose $\gamma\geq\beta$ to encourage disentanglement in the latent space.

The training process of WTC-WAE is similar to WTC-VAE with one major difference: two separate critics are needed to evaluate both the WTC term as well as the Wasserstein-$1$ distance between the prior and the factorized posterior. The training algorithm is outlined in Appendix %\ref{app:algo_wtc_wae}.
A.

\section{Evaluating disentanglement with Wasserstein dependency gap}
To evaluate disentanglement, previous work focuses on quantifying the statistical relations between the learned representation and the ground truth factors \cite{chen2018isolating,eastwood2018framework,higgins2017beta,kim2018disentangling,kumar2017variational,ridgeway2018learning}. For example, FactorVAE score \cite{kim2018disentangling} uses a majority vote classifier to predict a fixed factor of variation; MIG \cite{chen2018isolating} measures for each factor of variation the normalized gap in mutual information between the highest and second highest coordinate in the latent representations; Modularity \cite{ridgeway2018learning} measures whether a single latent dimension corresponds to at most one factor of variation.

It has been shown that any high-confidence lower bound on the mutual information requires sample size exponential in the mutual information \cite{ozair2019wasserstein}. Therefore, estimate MIG requires exponentially large sample size with respect to the highest mutual information between factors of variation and latent representations. As this limitation comes from the use of KL divergence in the mutual information definition, it motivates us to use Wasserstein dependency measure \cite{ozair2019wasserstein} as a replacement. 
Denote $v_k$ as the generating factor of interest, the Wasserstein dependency between $v_k$ and $z_i$ is defined as:
\begin{align}
    I_{\mathcal{W}}(z_i, v_k) = \mathbb{E}_{p(v_k)}[\mathcal{W}_1(q(z_i|v_k), q(z_i))]
\end{align}

The new metric which we refer to as Wasserstein dependency gap (WDG) can be written as:
\begin{align}
\frac{1}{K} \sum_{k=1}^{K} \left(I_{\mathcal{W}}\left(z_{i^{(k)}}, v_{k}\right)-\max _{i \neq i(k)} I_{\mathcal{W}}\left(z_{i} ; v_{k}\right)\right)
\end{align}
where $i^{(k)}=\operatorname{argmax}_{i} I_{\mathcal{W}}\left(z_{i} , v_{k}\right)$. Fortunately, as both $q(z_i|v_k)$ and $q(z_i)$ are uni-variate, we can calculate the term efficiently with empirical cumulative distribution functions from both distributions. Compared to MIG, WDG has the benefits of being easy to estimate and at the same time robust to the random samples.

There are several other metrics proposed for disentanglement such as the $\beta$-VAE metric \cite{higgins2017beta} and the SAP score \cite{kumar2017variational}. \citet{locatello2019challenging} show that these metrics are more or less correlated with FactorVAE score and MIG.

\begin{table*}[t]
    %\small
    
    \centering
    
    \begin{subtable}[t]{\textwidth}
    \centering
    \begin{tabular}{lcccc}
        \toprule
         \textbf{Model} & \bf FactorVAE score & \bf WDG & \bf Modularity & \bf Reconstruction \\
         \midrule
         $\beta$-VAE & 0.919 (0.042) & 0.023 (0.008) & \underline{0.900} (0.005) & 1412.5 (23.5) \\
         FactorVAE & \underline{0.930} (0.024) & 0.030 (0.013) & 0.893 (0.004) & 1462.0 (25.3) \\
         $\beta$-TCVAE & 0.909 (0.037) & 0.026 (0.012) & 0.875 (0.005) & 1452.5 (23.7)\\
         WTC-VAE (Ours) & \textbf{0.939} (0.032) & \underline{0.034} (0.013) & 0.897 (0.004) & \underline{1396.8} (25.6)\\
         \midrule
         WAE & 0.839 (0.048) & 0.018 (0.007) & \textbf{0.910} (0.005) & \textbf{1394.7} (23.5)\\
         WTC-WAE (Ours) & 0.907 (0.034) & \textbf{0.054} (0.011) & 0.902 (0.004) & 1402.2 (25.7)\\
        \bottomrule
    \end{tabular}
    \caption{Evaluation results on \textit{Cars3D} data set.}
    \vspace{1em}
    \end{subtable}

    \begin{subtable}[t]{\textwidth}
    \centering
    \begin{tabular}{lcccc}
        \toprule
         \textbf{Model} & \bf FactorVAE score & \bf WDG & \bf Modularity & \bf Reconstruction \\
         \midrule
         $\beta$-VAE & 0.607 (0.059) & 0.051 (0.005) & \underline{0.717} (0.010) & 45.5 (0.3) \\
         FactorVAE & \underline{0.738} (0.048) & 0.056 (0.006) & 0.667 (0.009) & 22.6 (0.7)\\
         $\beta$-TCVAE & \textbf{0.755} (0.042) & 0.056 (0.008) & 0.696 (0.007) & 27.2 (0.8)\\
         WTC-VAE (Ours) & 0.685 (0.070) & \underline{0.059} (0.006) & 0.679 (0.010) & 13.8 (0.5) \\
         \midrule
         WAE & 0.514 (0.046) & 0.024 (0.006) & 0.671 (0.010) & \textbf{9.7} (0.4)\\
         WTC-WAE (Ours) & 0.672 (0.031) & \textbf{0.088} (0.007) & \textbf{0.718} (0.009) & \underline{11.4} (0.5)\\
        \bottomrule
    \end{tabular}
    \caption{Evaluation results on \textit{dSprites} data set.\\}
    \vspace{1em}
    \end{subtable}

    \begin{subtable}[t]{\textwidth}
    \centering
    \begin{tabular}{lcccc}
        \toprule
         \textbf{Model} & \bf FactorVAE score & \bf WDG & \bf Modularity & \bf Reconstruction \\
         \midrule
         $\beta$-VAE & \underline{0.968} (0.018) & \textbf{0.138} (0.006) & \underline{0.790} (0.013) & 3553.7 (24.7) \\
         FactorVAE & 0.957 (0.026) & 0.104 (0.007) & \textbf{0.811} (0.014) & 3548.3 (24.7) \\
         $\beta$-TCVAE & 0.939 (0.036)  & 0.079 (0.006) & 0.659 (0.012) & 3521.6 (24.6)\\
         WTC-VAE (Ours) & \textbf{0.988} (0.030) & 0.087 (0.006) & 0.707 (0.014) & \underline{3520.7} (24.6)\\
         \midrule
         WAE & 0.671 (0.057) & 0.022 (0.005) & 0.606 (0.016) & \textbf{3517.4} (24.6)\\
         WTC-WAE (Ours) & 0.892 (0.027) & \underline{0.116} (0.006) & 0.774 (0.015) & 3532.4 (24.6)\\
        \bottomrule
    \end{tabular}
    \caption{Evaluation results on \textit{Shapes3D} data set.}
    \end{subtable}
    \caption{Mean (Std) of disentanglement scores and their corresponding reconstruction errors for all methods. Note that the higher the better for FactorVAE score, WDG, and Modularity while the lower the better for reconstruction. Best mean values are highlighted in bold. Second best mean values are underlined.}
    \label{tab:results}
\end{table*}

\section{Experiments}
\subsection{Experimental settings}
We first introduce the data sets, baselines, some modeling choices and evaluation metrics.
\paragraph{Data sets} We conduct quantitative experiments on three data sets with ground truth factors (i.e. factors of variation that control the generation process): 
\begin{itemize}
    \item \textit{Cars3D} \cite{reed2015deep}: 17,568 RGB $64\times 64\times 3$ images of car renderings with three ground truth factors of variation
    \item \textit{dSprites} \cite{higgins2017beta}: 737,280 gray-scale $64\times 64$ images of sprites
    \item \textit{Shapes3D} \cite{kim2018disentangling}: 480,000 RGB $64\times 64\times 3$ images of 3D shapes with 6 different ground truth factors
\end{itemize}
We also experiment on a cropped version of \textit{CelebA} \cite{liu2015faceattributes} to qualitatively examine the results. A summary of the data sets and their corresponding ground truth factors can be found in Appendix %\ref{app:data}.
B.

\paragraph{Baselines} For VAE-based models, we compare with $\beta$-VAE \cite{higgins2017beta}, FactorVAE \cite{kim2018disentangling}, and $\beta$-TCVAE \cite{chen2018isolating} on their disentanglement performances as well as reconstruction abilities. We also include the comparison of WAE and WTC-WAE. Note that for WAE-based models, we choose to use Wasserstein distance to measure distribution discrepancies. We do not compare with other options such as JS divergence and MMD.

\paragraph{Inductive biases} 
To fairly compare all approaches, we use the same convolutional architecture in all the autoencoding components. All methods, including WAE-based approaches, use a Gaussian encoder, a Bernoulli decoder and latent dimension fixed to 10. We use Adam optimizers \cite{kingma2014adam} with the same hyper-parameter setting across all experiments. These settings are standard choices in prior work \cite{chen2018isolating,higgins2017beta,kim2018disentangling,locatello2019challenging}. The detailed settings can be found in Appendix %\ref{app:exp_settings}.
C.

\paragraph{Metrics} To quantitatively evaluate the model performances, we consider three disentanglement metrics: FactorVAE score, Modularity, and Wasserstein dependency gap (WDG). We also evaluate models' abilities to achieve these disentanglement scores without large compromises in reconstructions.

Our implementation is based on PyTorch v1.0.0\footnote{https://pytorch.org/} and will be made available.

\begin{figure*}[t]
\centering
\begin{minipage}[b]{0.32\textwidth}
\centering
\includegraphics[width=\textwidth]{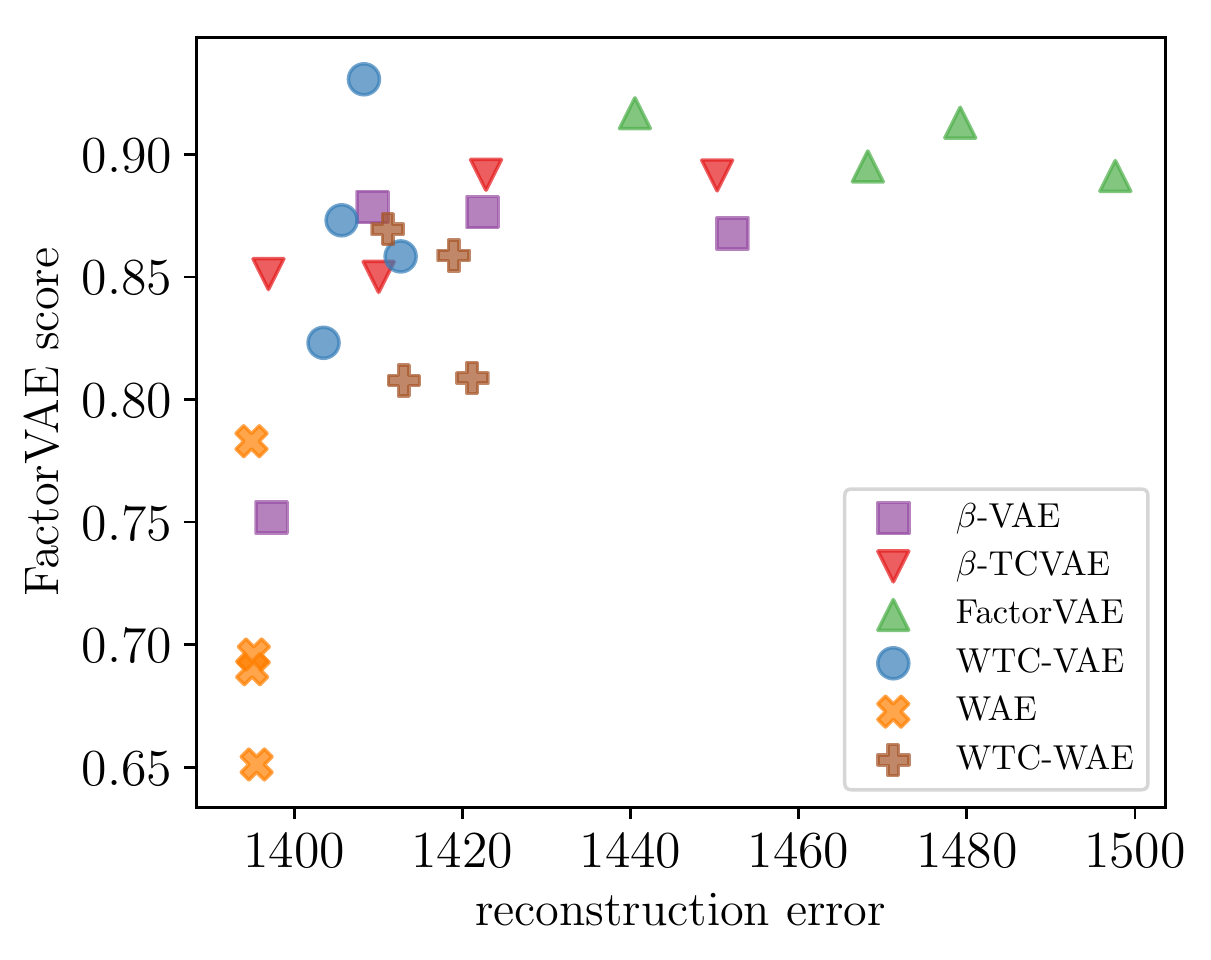}
\\
(a)
\end{minipage}
\hfill
\begin{minipage}[b]{0.32\textwidth}
\centering
\includegraphics[width=\textwidth]{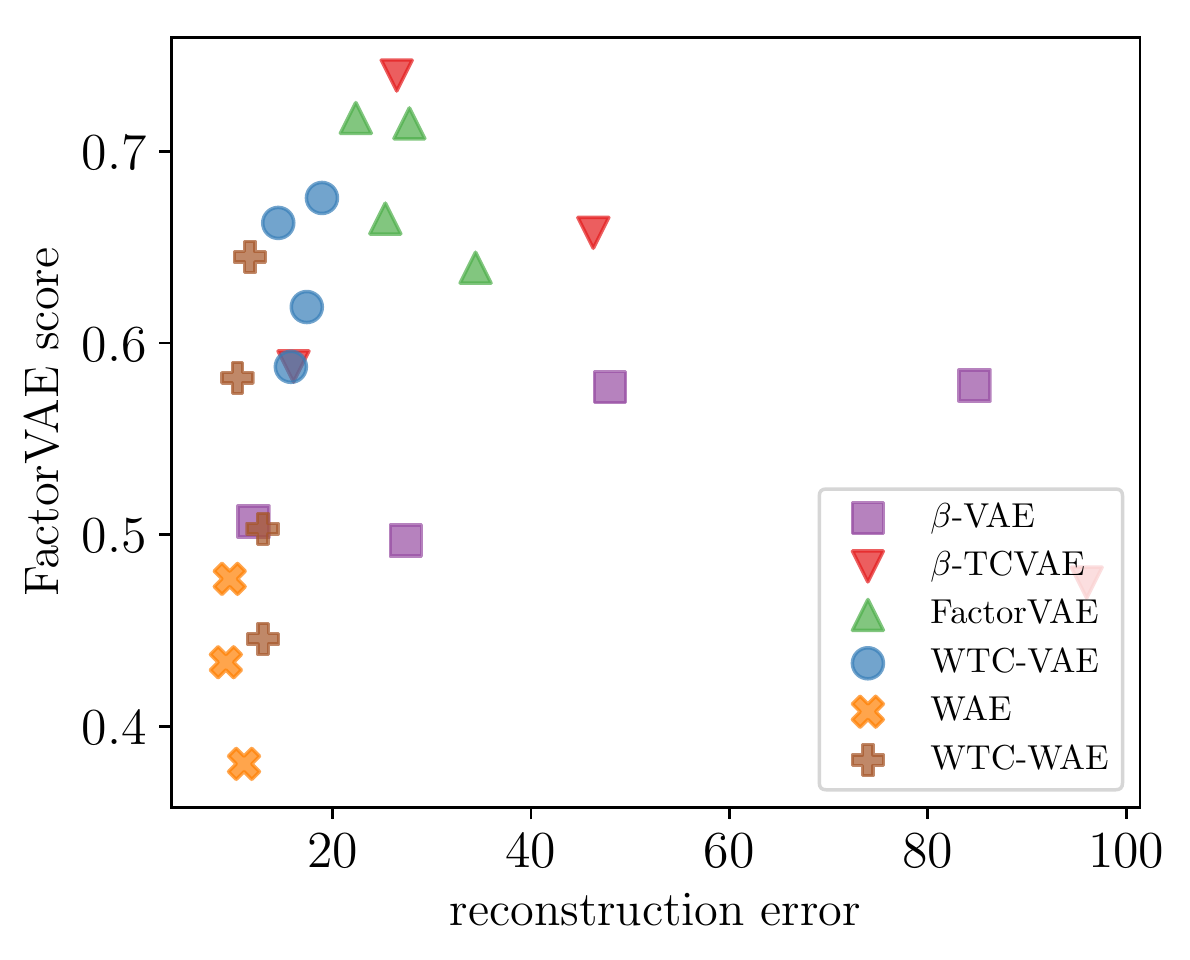}
\\
(b)
\end{minipage}
\hfill
\begin{minipage}[b]{0.32\textwidth}
\centering
\includegraphics[width=\textwidth]{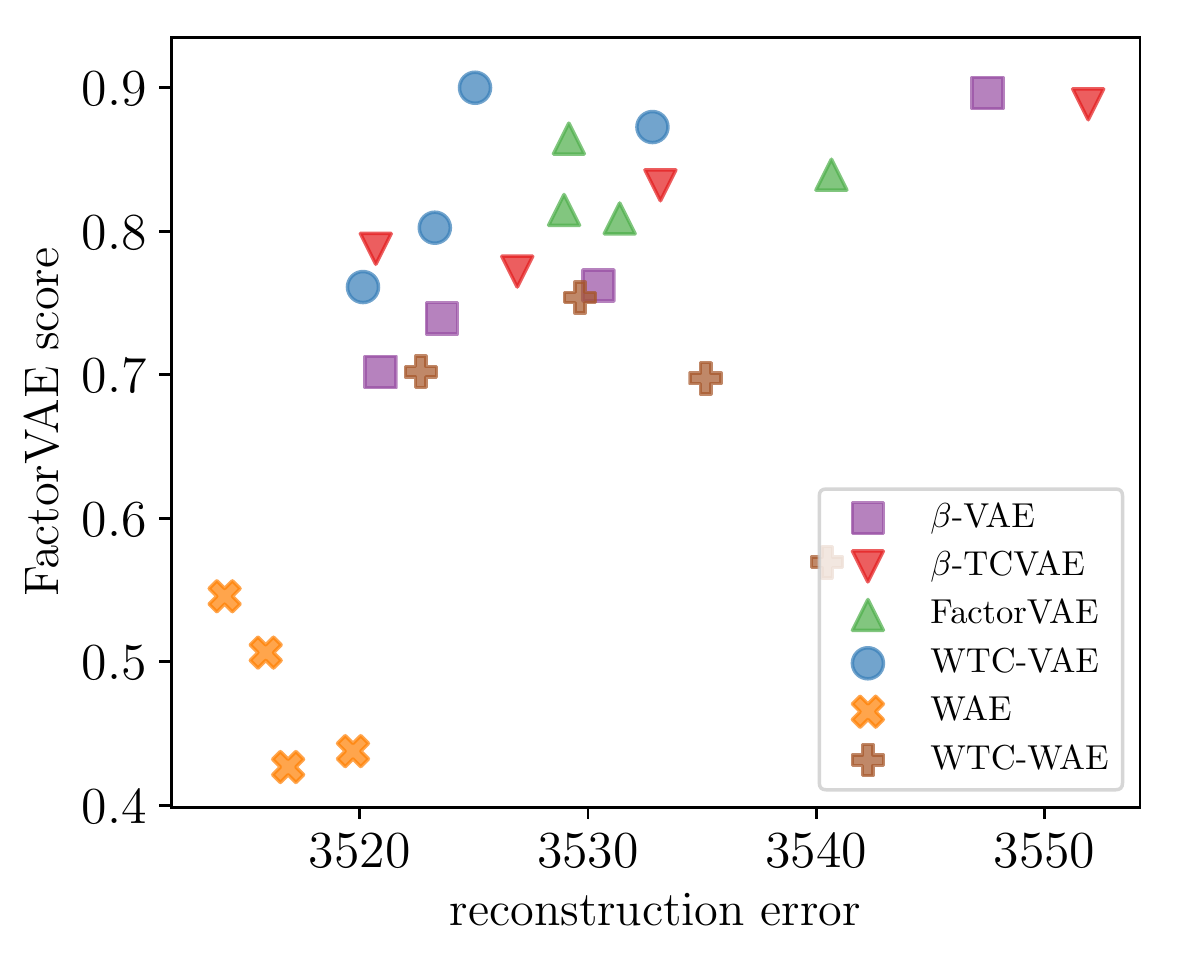}
\\
(c)
\end{minipage}
\caption{Scatter plots of FactorVAE scores against reconstruction errors on (a) Cars3D, (b) dSprites, and (c) Shapes3D. Each point is obtained for one specific regularization strength averaged over five random seeds. Upper-left is better. WTC-VAE consistently performs on par with other models while maintaining lower reconstruction errors.}
    \label{fig:scatter}
\end{figure*}

\subsection{Disentanglement performances}
It has been shown in previous work \cite{chen2018isolating,locatello2019challenging} that unsupervised disentangled representation learning typically has high variance with respect to the defined metrics as the ground truth factors are unavailable during training. To account for the variance, we train all methods with different random seeds. For each approach, four regularization strengths are chosen based on the reported values in corresponding previous work. We report the mean and standard deviation of the disentanglement scores as well as reconstruction errors for each approach with their best regularization strength setting. The results are listed in Table \ref{tab:results}.

We can observe that WTC-VAE performs mostly on par with respect to the three disentanglement metrics while maintaining the lowest reconstruction error across all three data sets compared to other VAE-based approaches. WTC-WAE also shows consistent performance gains over its baseline WAE.

\subsection{Trade-off between disentanglement and reconstruction}
In general, when optimizing the model to learn increasingly disentangled representations, the reconstruction error is expected to increase as well because of different weightings in the objective function. As observed in Table \ref{tab:results}, WTC-VAE has a better balance between disentanglement and reconstruction performances. Figure \ref{fig:scatter} shows scatter plots of FactorVAE scores against reconstruction errors on all three data sets. Each point represents the average performance of a model trained with the same regularization strength over different random seeds. WAE has the lowest reconstruction errors but performs poorly on disentanglement. WTC-VAE consistently positions on the upper left corner which indicates higher disentanglement scores and lower reconstruction errors. Similar plots for WDG can be found in Appendix %\ref{app:further_results}.
E.

% \begin{figure*}[t]
% \centering
% \begin{minipage}[b]{0.32\textwidth}
% \centering
% \includegraphics[width=\textwidth]{figs/violin_wtc_wmig_cars3d.pdf}
% \\
% (a)
% \end{minipage}
% \hfill
% \begin{minipage}[b]{0.32\textwidth}
% \centering
% \includegraphics[width=\textwidth]{figs/violin_wtc_wmig_dsprites.pdf}
% \\
% (b)
% \end{minipage}
% \hfill
% \begin{minipage}[b]{0.32\textwidth}
% \centering
% \includegraphics[width=\textwidth]{figs/violin_wtc_wmig_shapes3d.pdf}
% \\
% (c)
% \end{minipage}
% \caption{Violin plots of WDG evaluations for WTC-VAE with different regularization strengths on (a) Cars3D, (b) dSprites, and (c) Shapes3D. There is a correlation between WDG and regularization strength on two of the three data sets.}
%     \label{fig:violin_wtc}
% \end{figure*}
\begin{figure}[t]
\centering

\includegraphics[width=0.36\textwidth]{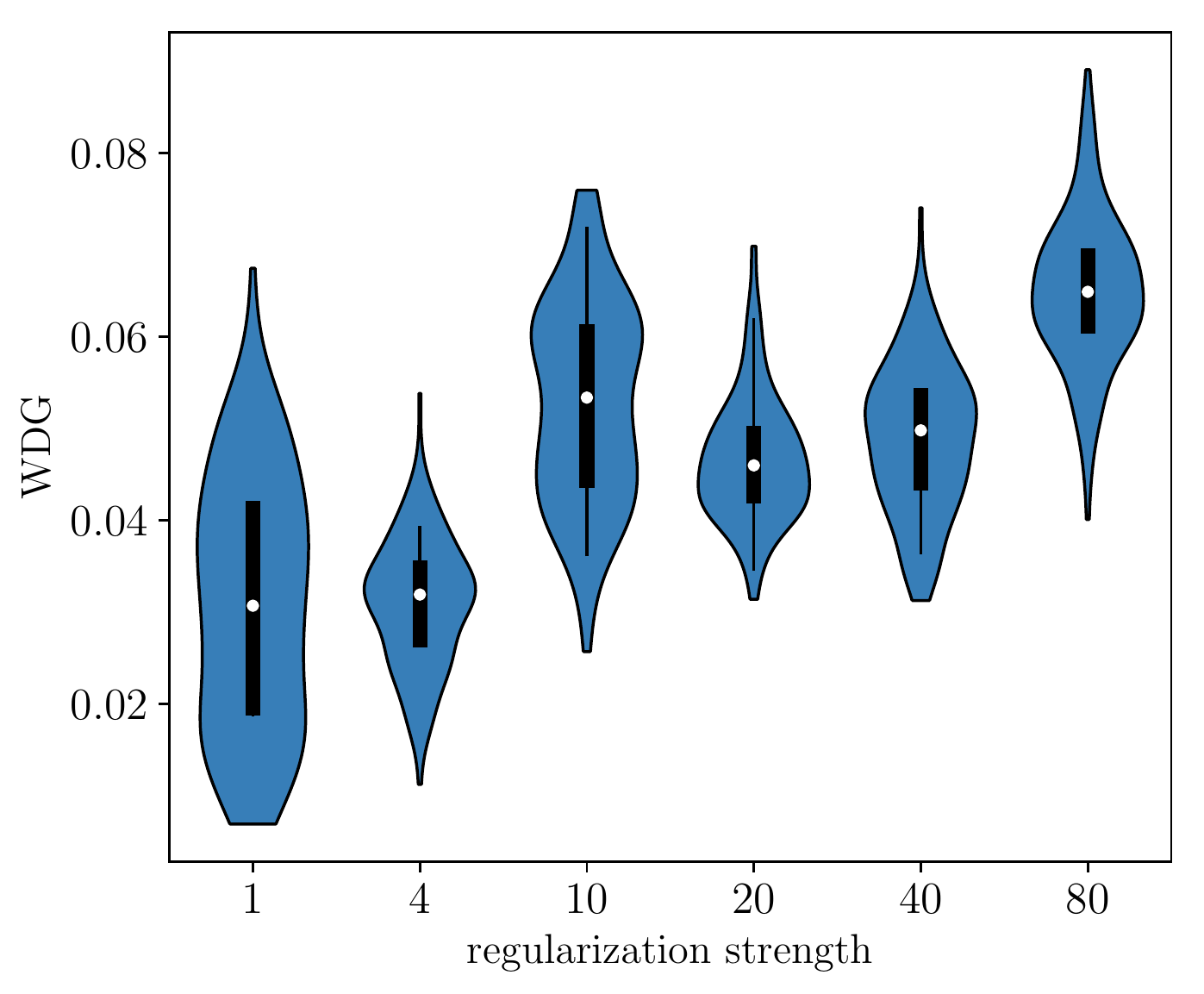}

\caption{Violin plots of WDG evaluations for WTC-VAE with different regularization strengths on dSprites. There is a correlation between WDG and regularization strength.}
    \label{fig:violin_wtc}
\end{figure}

\subsection{Effect of regularization strength}
To further investigate the influences of regularization strength of the WTC term on disentanglement performances, we conduct experiments using WTC-VAE with six different regularization strength $\gamma\in\{1,4,10,20,40,80\}$. Again, each setting includes multiple runs with different random seeds. 

Figure \ref{fig:violin_wtc} shows violin plots of WDG evaluations for WTC-VAE with different regularization strengths. For Cars3D, regularization strength seems to have minor influences on the disentanglement performances. On the other hand, for dSprites and Shapes3D, it becomes obvious that WDG rises with larger regularization strengths. It is also worth noting that the worst performance with overall better hyper-parameter settings can be worse than the best performance with a regular setup. Similar results are observed with FactorVAE scores and can be found in Appendix %\ref{app:further_results}.
E.

\begin{figure*}[h]
\centering
\begin{minipage}[b]{0.24\textwidth}
\centering
\includegraphics[width=\textwidth]{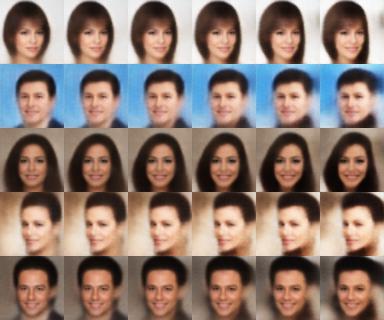}
\\
(a) Curly hair
\end{minipage}
\hfill
\begin{minipage}[b]{0.24\textwidth}
\centering
\includegraphics[width=\textwidth]{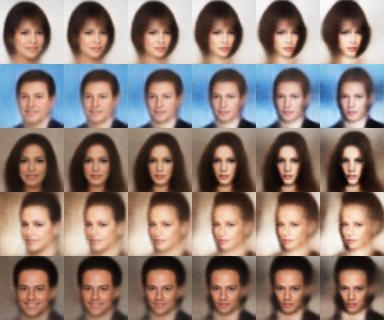}
\\
(b) Face width
\end{minipage}
\hfill
\begin{minipage}[b]{0.24\textwidth}
\centering
\includegraphics[width=\textwidth]{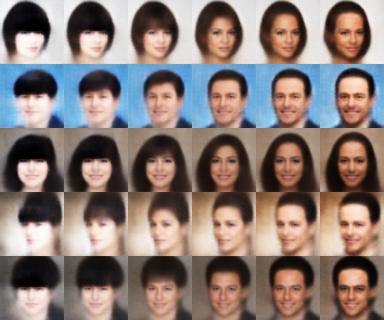}
\\
(c) Baldness
\end{minipage}
\hfill
\begin{minipage}[b]{0.24\textwidth}
\centering
\includegraphics[width=\textwidth]{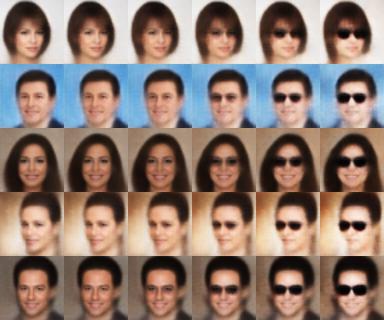}
\\
(d) Glasses
\end{minipage}
\caption{Latent space traversal of WTC-VAE on CelebA. Four example factors are shown.}
    \label{fig:celeba_trav}
\end{figure*}

\subsection{Rank correlation among different metrics}
\begin{figure}[h!]
\centering
\begin{minipage}[b]{0.3\textwidth}
\centering
\includegraphics[width=\textwidth]{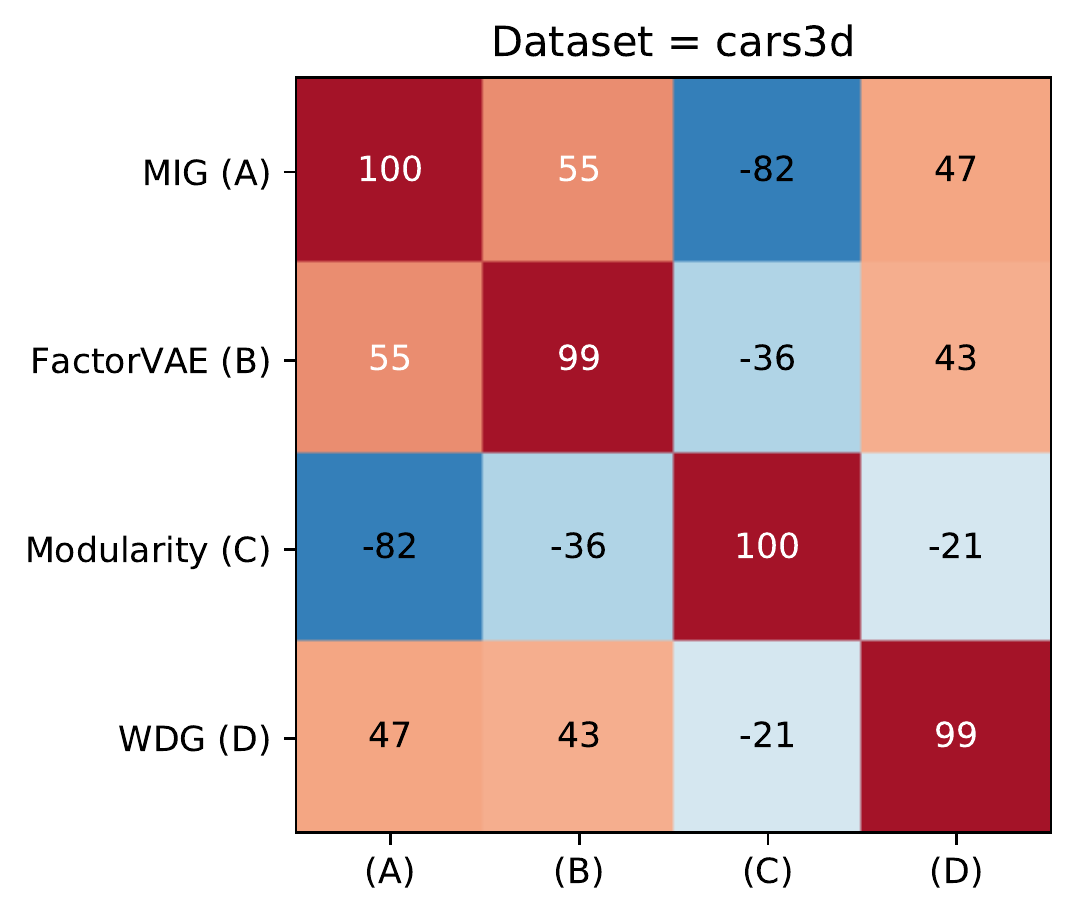}
\end{minipage}
\hfill
\begin{minipage}[b]{0.3\textwidth}
\centering
\includegraphics[width=\textwidth]{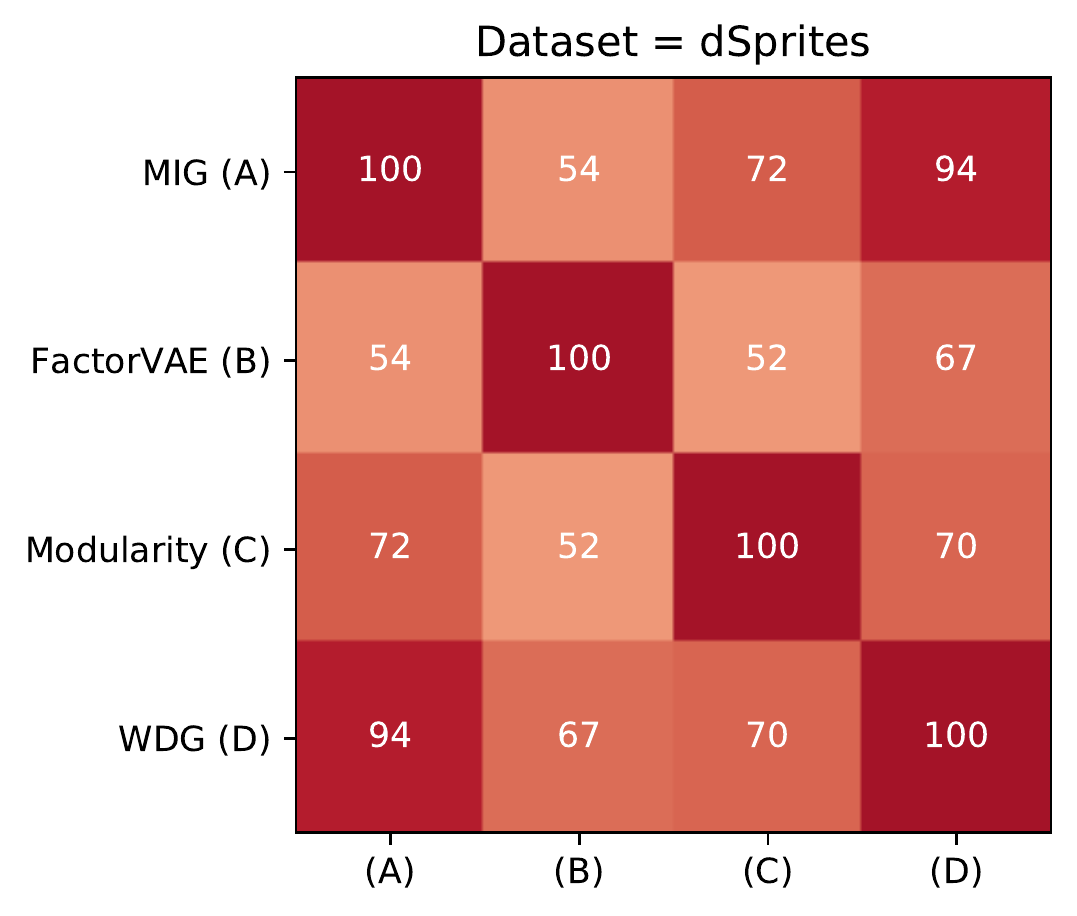}
\end{minipage}
\hfill
\begin{minipage}[b]{0.3\textwidth}
\centering
\includegraphics[width=\textwidth]{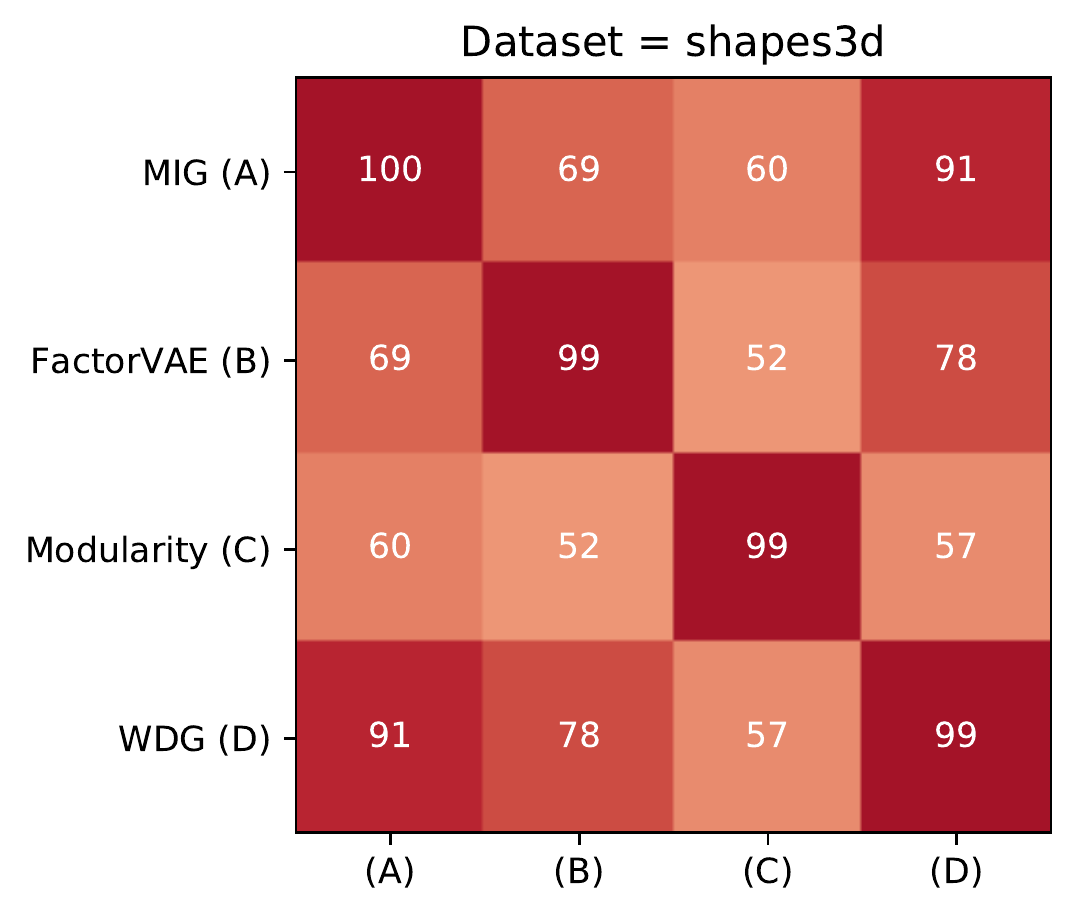}
\end{minipage}
\caption{Rank correlation between metrics for different data sets.}
    \label{fig:rankcorr}
\end{figure}

We measure the Pearson rank correlation among four metrics: FactorVAE score, MIG, WDG, and Modularity on the three labeled data sets. The results are shown in Figure \ref{fig:rankcorr}.

From the results we observe that WDG has high rank correlations with all three other metrics (except with Modularity on Cars3D).
\subsection{Qualitative results on CelebA}

To examine the quality of disentanglement learned with our proposed methods, we train WTC-VAE with $\gamma=40$ and latent dimension 20 on CelebA and traverse the latent space on different dimensions. We choose similar traversal ranges to \cite{chen2018isolating} and manually check the results. Note that we crop the original $218\times 178$ images to $178\times 178$ by taking the center regions and then down-sampling to $64\times 64$. The cropping method might be different in other related works. 

Figure \ref{fig:celeba_trav} shows some example factors of variation discovered by WTC-VAE. Interestingly, it discovers some subtle factors such as curly hair. We can also see that the learned disentangled representations are not perfect. For example, the factor of baldness is clearly entangled with the factor of lightness in the given examples.

\section{Conclusion}
In this paper, we introduce a new Wasserstein distance-based total correlation and leverage it to learn disentangled representations from images. Compared to total correlation, it has the benefit of being robust to estimations with random samples. With the same motivation, we propose Wasserstein dependency gap, a new metric to measure disentanglement. Our experimental results show that the proposed approach has comparable if not better disentanglement performances while achieving the lowest reconstruction errors on three vision data sets. Qualitative experiments on CelebA data set confirm that the proposed approach is able to discover meaningful factors of variation describing the images.

Disentangled representation learning remains a difficult problem. Introducing formal and applicable definitions of the problem, consistent learning framework for inputs beyond the vision domain, as well as more reliable evaluation criteria in the unsupervised settings are important directions for future work in this area.

\bibliographystyle{named}

\bibliography{ijcai20}

\appendix
\section*{Appendix}

\subsection*{A. Training algorithm for WTC-WAE}
\label{app:algo_wtc_wae}
Training algorithm for WTC-WAE is outlined in Algorithm \ref{alg:wtc_wae}. The main difference from WTC-VAE is the usage of two different critics $f, g$.
\begin{algorithm}[h]
\small
\DontPrintSemicolon
\IncMargin{1in}
\KwIn{Regularization coefficient $\beta, \gamma$, learning rate $\alpha$, batch size $B$.}
\KwIn{initial critic parameters $\theta,\phi$, initial VAE parameter $w$}
\While{$(\theta, \phi, w)$ not converged}{
sample batch $(x^{(i)})_{i\in\mathcal{B}}$ of size $B$\;
sample $z^{(i)}\sim q_w(z|x^{(i)}), \forall i\in \mathcal{B}$\;
$(\bar{z}^{(i)})_{i\in\mathcal{B}}\leftarrow$\texttt{permute\_dims}($(z_w^{(i)})_{i\in\mathcal{B}}$)\;
sample $z'^{(i)}\sim p(z), \forall i\in \mathcal{B}$\;
$\theta\leftarrow\theta+\alpha\cdot\nabla_\theta\frac{1}{B}\sum_{i\in\mathcal{B}}[f_\theta(z^{(i)})-f_\theta(\bar{z}^{(i)})]$\;
$\phi\leftarrow\phi+\alpha\cdot\nabla_\phi\frac{1}{B}\sum_{i\in\mathcal{B}}[g_\phi(\bar{z}^{(i)})-g_\phi(z'^{(i)})]$\;
$w\leftarrow w-\alpha\cdot\nabla_w\frac{1}{B}\sum_{i\in\mathcal{B}}[\beta(g_\phi(\bar{z}^{(i)})-g_\phi(z'^{(i)}))+\gamma(f_\theta(z^{(i)})-f_\theta(\bar{z}^{(i)}))-\log{p_w(x^{(i)}|z^{(i)})}]$
}
\caption{Training of WTC WAE}
\label{alg:wtc_wae}
\end{algorithm}

\subsection*{B. Summary of data sets}
\label{app:data}
Data sets and ground truth factors are summarized in Table \ref{tab:data}.
\begin{table}[h]
    \small
    \medskip
    \centering
    \begin{tabular}{lll}
    \toprule
        \bf Data set & \bf Size & \bf Ground truth factors \\
        \midrule
        \multirow{3}{*}{Cars3D}  & \multirow{3}{*}{17,568} & identity (183), azimuth (24), \\
         &  & elevation (4)\\
         & & \\
        \multirow{3}{*}{dSprites}  & \multirow{3}{*}{737,280} & scale (6), rotation (40),  \\
         &  & posX (32), posY (32)\\
         & & \\
        \multirow{3}{*}{Shapes3D}  & \multirow{3}{*}{480,000} & shape (4), scale (8), orientation (15), \\
        & & floor color (10), wall color (10), \\
        & & object color (10)\\
        \multirow{3}{*}{CelebA}  & \multirow{3}{*}{202,599} & \multirow{3}{*}{-} \\
         &  & \\
        \bottomrule
    \end{tabular}
    \caption{Data size and ground truth factors of the data sets used in the experiments. Number of different values for each factor is given in parentheses.}
    \label{tab:data}
\end{table}

\begin{table}
    \small
    \medskip
    \centering
    \begin{tabular}{l}
    \toprule
         \bf Encoder \\
         \midrule
         Input: 64$\times$ 64$\times$ num\_channels  \\
         4$\times$ 4 conv, 32 ReLU, stride 2 \\
         4$\times$ 4 conv, 32 ReLU, stride 2 \\
         4$\times$ 4 conv, 64 ReLU, stride 2\\
         4$\times$ 4 conv, 64 ReLU, stride 2 \\
         FC 256, FC 2$\times$ 10 \\
         \midrule
         \bf Decoder \\
         \midrule
         Input: $\mathbb{R}^{10}$ \\
         FC, 256 ReLU \\
         FC, 4$\times$ 4$\times$ 64 ReLU \\
         4$\times$ 4 upconv, 64 ReLU, stride 2\\
         4$\times$ 4 upconv, 32 ReLU, stride 2\\
         4$\times$ 4 upconv, 32 ReLU, stride 2\\
         4$\times$ 4 upconv, num\_channels, stride 2\\
         \midrule
         \bf Critic \\
         \midrule
         Input: $\mathbb{R}^{10}$\\
         FC, 256 ReLU \\
         FC, 256 ReLU \\
         FC, 256 ReLU \\
         FC, 1 \\
         \bottomrule
    \end{tabular}
    \caption{Encoder, decoder and critic architectures.}
    \label{tab:arch}
\end{table}

\begin{table}
    \small
    \centering
    \begin{subtable}[h]{0.45\textwidth}
    \centering
    \begin{tabular}{lll}
    \toprule
         \bf Model & \bf Parameter & \bf Values\\
         \midrule
         $\beta$-VAE & $\beta$ & [1, 4, 8, 16] \\
         FactorVAE & $\gamma$ & [10, 20, 40, 80] \\
         $\beta$-TCVAE & $\beta$ & [1, 4, 8, 16] \\
         WTC-VAE & $\gamma$ & [10, 20, 40, 80] \\
         WAE & $\beta$ & [1, 4, 8, 16] \\
         WTC-WAE & $\gamma$ & [10, 20, 40, 80] \\
         \bottomrule
    \end{tabular}
    \caption{Regularization strength hyper-parameter settings.}
    \vspace{1em}
    \end{subtable}

    \begin{subtable}[h]{0.45\textwidth}
    \centering
    \begin{tabular}{ll}
    \toprule
         \bf Parameter & \bf Values\\
         \midrule
         Batch size & 64 \\
         Optimizer & Adam \\
         Adam: $\beta_1$ & 0.9 \\
         Adam: $\beta_2$ & 0.999 \\
         Adam: $\epsilon$ & 1e-8 \\
         Adam: learning rate & 0.0001 \\
         Training steps & 300000 \\
         \bottomrule
    \end{tabular}
    \caption{Optimization hyper-parameter settings.}
    \end{subtable}
    \caption{Hyper-parameter settings.}
    \label{tab:hparams}
\end{table}

\subsection*{C. Detailed experimental settings}
\label{app:exp_settings}
\paragraph{Model architecture} We use the same convolutional neural network architectures across all experiments. For the critics, we use multi-layer perceptrons (MLPs) with ReLU. The hyper-parameters for the architecture are listed in Table \ref{tab:arch}.

\paragraph{Regularization strength} For each model being compared in the experiments, we choose four different regularization strengths. These hyper-parameters are listed in Table \ref{tab:hparams}(a).

\begin{figure*}[h]
\centering
\begin{minipage}[b]{0.32\textwidth}
\centering
\includegraphics[width=\textwidth]{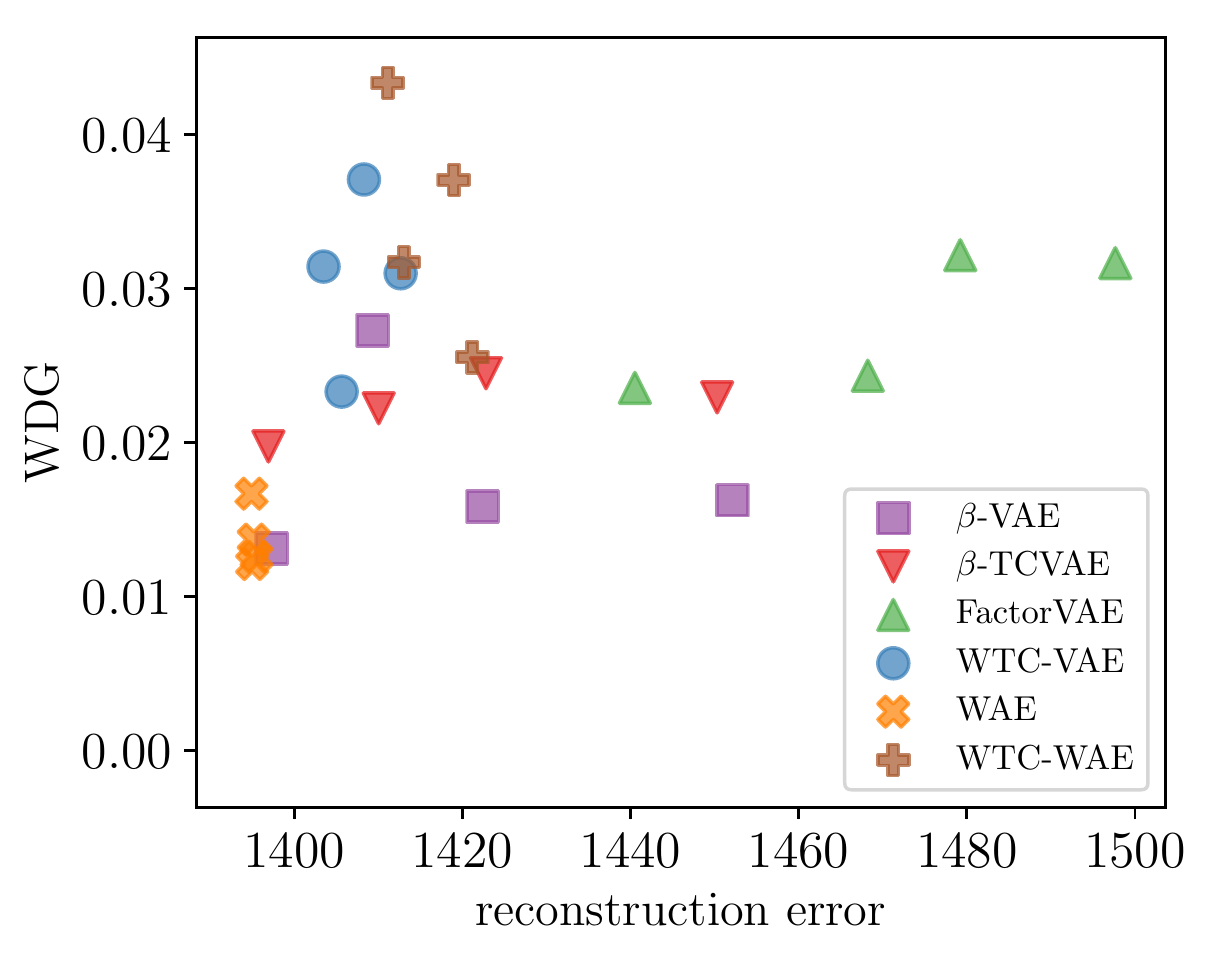}
\\
(a)
\end{minipage}
\hfill
\begin{minipage}[b]{0.32\textwidth}
\centering
\includegraphics[width=\textwidth]{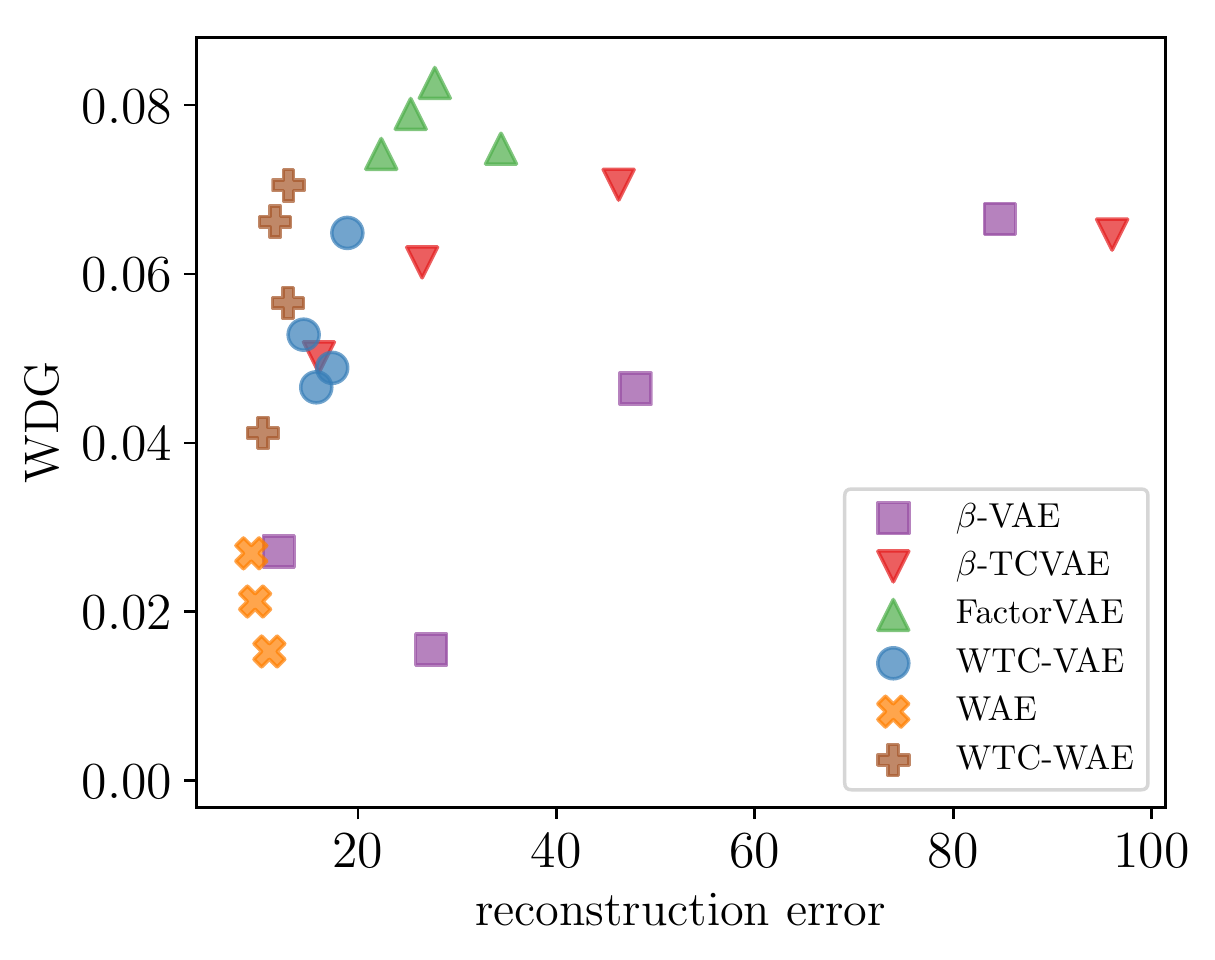}
\\
(b)
\end{minipage}
\hfill
\begin{minipage}[b]{0.32\textwidth}
\centering
\includegraphics[width=\textwidth]{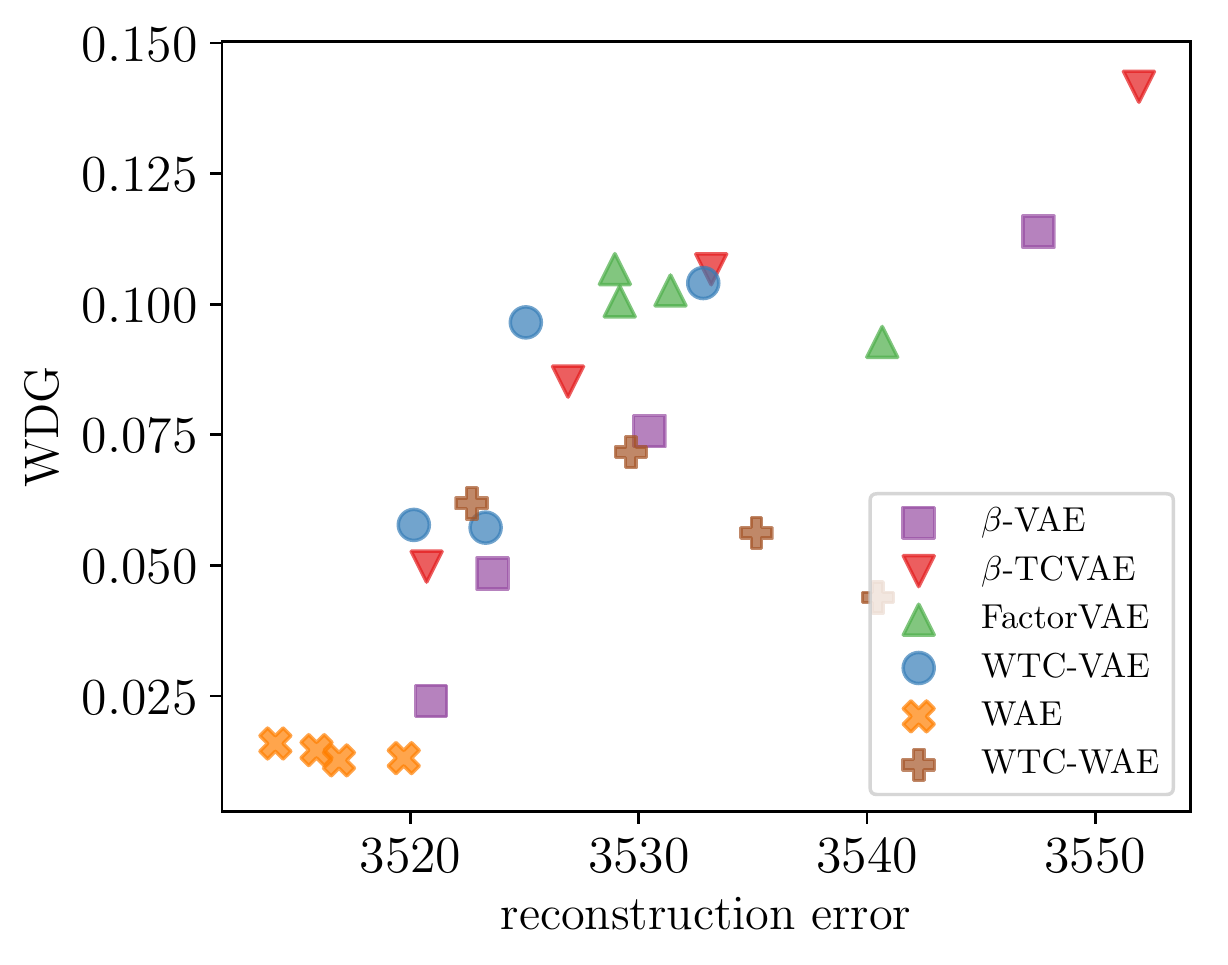}
\\
(c)
\end{minipage}
\caption{Scatter plots of WDG against reconstruction errors on (a) Cars3D, (b) dSprites, and (c) Shapes3D. Each point is obtained for one specific regularization strength averaged over five random seeds. Upper-left is better.}
    \label{fig:scatter_wdg}
\end{figure*}

\paragraph{Optimization} We use Adam optimizer %\cite{kingma2014adam} 
with default parameter settings except for the learning rate. The settings are listed in Table \ref{tab:hparams}(b).

\subsection*{D. Implementation of metrics}
\label{app:metrics}
All metrics use the mean values of the latent representations. Due to the randomness in the evaluation, we run each evaluation 50 times.

\paragraph{FactorVAE score} Firstly, we estimate the variance of each latent dimension by embedding 12800 random samples from the data and exclude collapsed dimensions with variance smaller than 0.05. Second, we sample a batch of points with a randomly fixed factor and evaluate the variance of each latent dimension. A vote is obtained by associating the latent dimension with the lowest normalized variance with the fixed factor. We train on 10000 votes and evaluate on 5000 votes.

\paragraph{Wasserstein dependency gap} First, we embed 12800 random examples into latent space and normalize the latent variables to be scale invariant. Second, for each possible value of the factor of interest, we calculate the Wasserstein distance between a specific latent dimension and another random value from the same mini-batch. Third, Wasserstein dependency (WD) between a latent dimension and a factor is taken to be the average over all distances calculated with unique values of the factor. WDG is then computed as the difference between the highest and second highest WD.

\subsection*{E. Further experimental results}
\label{app:further_results}
\begin{figure}[h!]
\centering
\begin{minipage}[b]{0.27\textwidth}
\centering
\includegraphics[width=\textwidth]{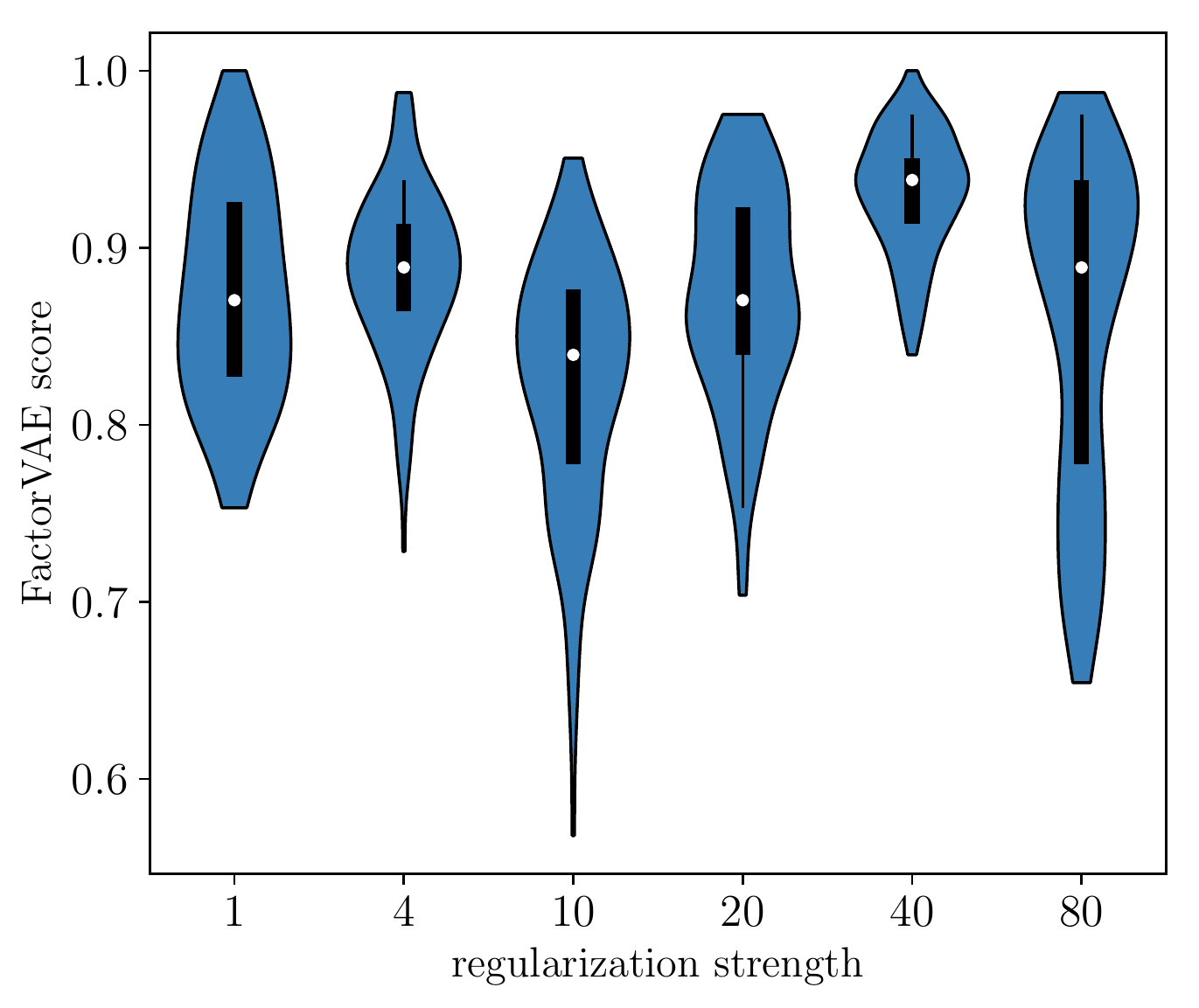}
\\
(a)
\end{minipage}
\hfill
\begin{minipage}[b]{0.27\textwidth}
\centering
\includegraphics[width=\textwidth]{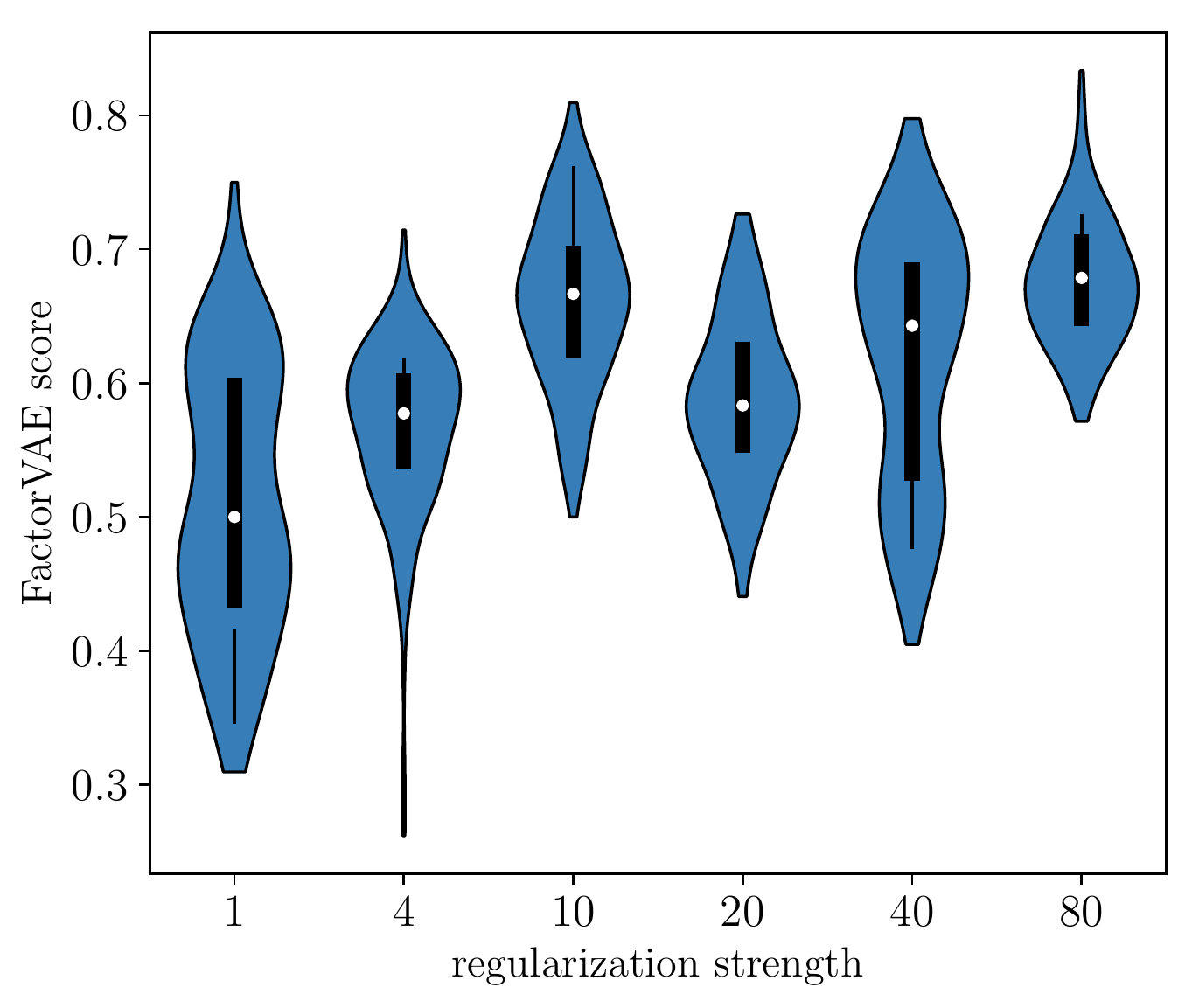}
\\
(b)
\end{minipage}
\hfill
\begin{minipage}[b]{0.27\textwidth}
\centering
\includegraphics[width=\textwidth]{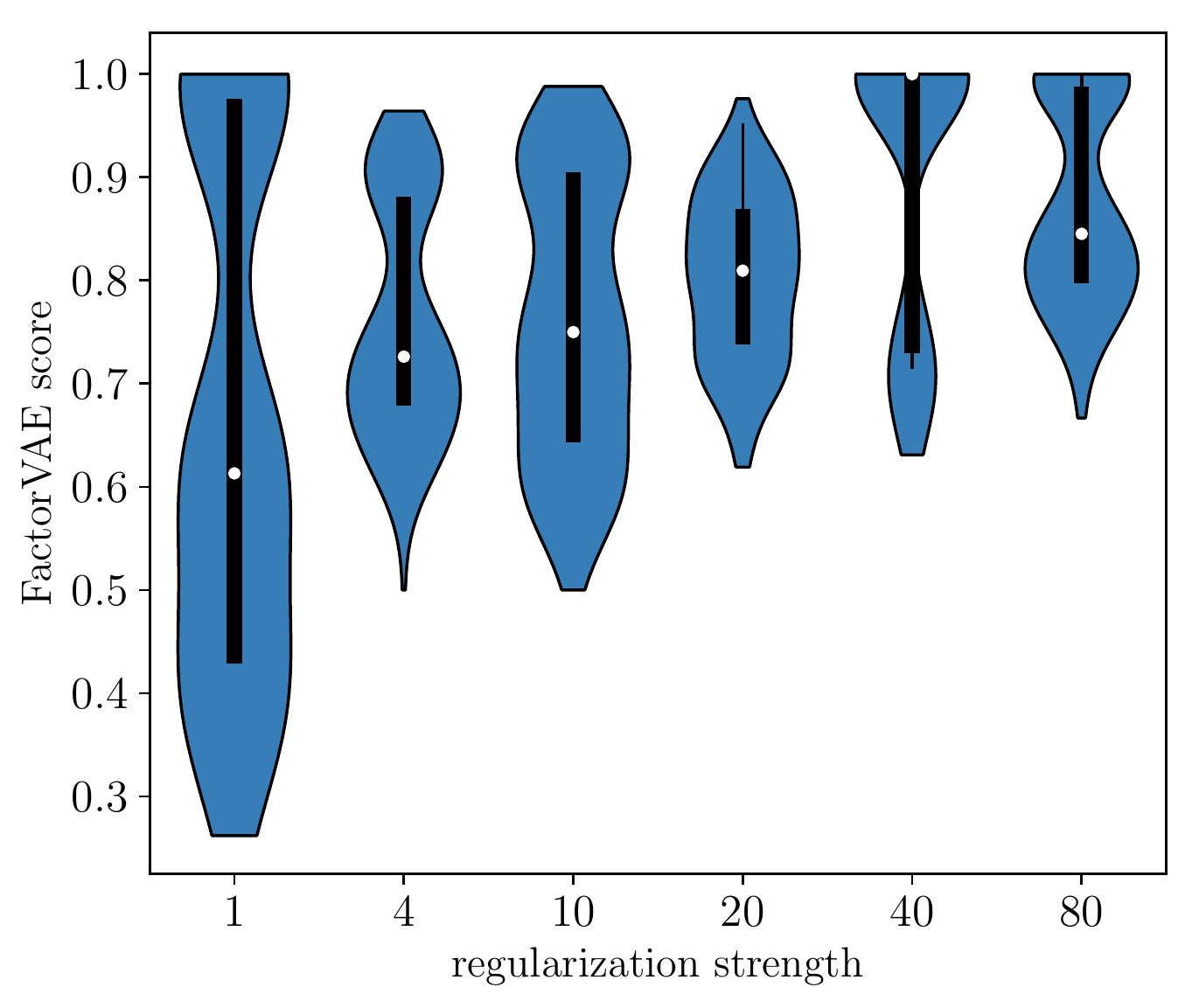}
\\
(c)
\end{minipage}
\caption{Violin plots of FactorVAE score evaluations for WTC-VAE with different regularization strengths on (a) Cars3D, (b) dSprites, and (c) Shapes3D.}
    \label{fig:violin_wtc_factor}
\end{figure}

Figure \ref{fig:scatter_wdg} shows the scatter plots of WDG against reconstruction errors on all three data sets. Figure \ref{fig:violin_wtc_factor} shows violin plots of FactorVAE scores for WTC-VAE with different regularization strengths.

\end{document}